\documentclass[conference]{IEEEtran}
\IEEEoverridecommandlockouts
\usepackage{cite}
\usepackage{amsmath,amssymb,amsfonts}
\usepackage{graphicx}
\usepackage{textcomp}
\usepackage{xcolor}
\def\BibTeX{{\rm B\kern-.05em{\sc i\kern-.025em b}\kern-.08em
    T\kern-.1667em\lower.7ex\hbox{E}\kern-.125emX}}

\usepackage{multirow}
\usepackage{algorithm}
\usepackage{algpseudocode}

\algdef{SE}[DOWHILE]{Do}{doWhile}{\algorithmicdo}[1]{\algorithmicwhile\ #1}%

\begin{document}

\newcommand{\networkNameNoSpace}{Dendron}

\title{\networkNameNoSpace: Enhancing Human Activity Recognition with On-Device TinyML Learning
}

\author{\IEEEauthorblockN{Hazem Hesham Yousef Shalby}
\IEEEauthorblockA{\textit{Politecnico di Milano}\\
Milan, Italy \\
hazemhesham.shalby@polimi.it}
\and
\IEEEauthorblockN{Manuel Roveri}
\IEEEauthorblockA{\textit{Politecnico di Milano}\\
Milan, Italy \\
manuel.roveri@polimi.it}
}

\maketitle

\begin{abstract}
Human activity recognition (HAR) is a research field that employs Machine Learning (ML) techniques to identify user activities.
Recent studies have prioritized the development of HAR solutions directly executed on wearable devices, enabling the on-device activity recognition. This approach is supported by the Tiny Machine Learning (TinyML) paradigm, which integrates ML within embedded devices with limited resources. However, existing approaches in the field lack in the capability for on-device learning of new HAR tasks, particularly when supervised data are scarce.
To address this limitation, our paper introduces \textit{\networkNameNoSpace}, a novel TinyML methodology designed to facilitate the on-device learning of new tasks for HAR, even in conditions of limited supervised data.
Experimental results on two public-available datasets and an off-the-shelf device (STM32-NUCLEO-F401RE) show the effectiveness and efficiency of the proposed solution. 

\end{abstract}


\section{Introduction}\label{sec:introduction}
Human activity recognition (HAR) is a research area focusing on developing systems that can automatically identify user activities (e.g., lying, standing, walking, or running) by using Machine Learning (ML) techniques.
In the last few years, research in the field focused on the development of HAR solutions operating on wearable devices, so as to support the on-device recognition of human activities directly on tiny devices \cite{bevilacqua_human_2019,wang_hierhar_2021, wang_towards_2018, banos_human_2013, cho_divide_2018, pu_hrhar_2023}.

This novel research field enhances the Tiny Machine Learning (TinyML) perspective, which integrates ML within embedded devices, constrained by limited memory, low computational power, and low power consumption \cite{tiny_org}.

Interestingly, existing works in the field suffer from the absence of an on-device adaptation and integration of new tasks to the HAR model at runtime with limited supervised data availability \cite{pavan_tybox_2023,ren_tinyol_2021,craighero_-device_2023, alzubaidi_survey_2023}. 
However, this is essential in on-device HAR, as we cannot collect and store large datasets directly on the device.

This paper aims to introduce a novel methodology, called \textit{\networkNameNoSpace}, that allows the HAR model to learn new tasks directly on the wearable device in supervised data scarcity conditions.
Specifically, the proposed architecture employs a hierarchical approach that, unlike the traditional methods that utilize a single model for HAR, takes into account the similarities between activities and decomposes the multi-class classification problem into multiple stages of sub-classification \cite{wang_hierhar_2021}.
For example, the first stage of the hierarchical classification predicts a general-activity category (e.g., walking vs running), while subsequent stages refine the prediction by identifying specific sub-activities (e.g., walking upstairs vs downstairs).
With its specifically-designed architecture, \textit{\networkNameNoSpace} efficiently and effectively supports the on-device learning of new tasks.
Differently from traditional hierarchical approaches (e.g., \cite{wang_hierhar_2021}) that require multiple training processes (i.e., one for each sub-task), \textit{\networkNameNoSpace} has a unique training process, hence simplifying and reducing the complexity of the training phase.
Additionally, the unified training process of \textit{\networkNameNoSpace} allows for the effective and efficient learning of tasks with limited data availability.

The experimental results show that \textit{\networkNameNoSpace} has superior performance in on-device learning compared to existing solutions, particularly under conditions of limited supervised data, as evidenced by experimental results on two publicly available datasets \cite{novac_uca-ehar_2022,wermter_human_2014} and in the porting on a resource-constrained device (STM32-NUCLEO-F401RE).
Additionally, \textit{\networkNameNoSpace} uses $5\times$ less memory, requires $2\times$ less computational load, and takes $2\times$ less time per inference compared to other hierarchical solutions.

Summing up, the novel contributions of this paper are as follows:
\begin{itemize}
    \item a novel hierarchical architecture that allows the efficient and effective on-device learning of new activities under conditions of supervised data scarcity;
    \item a novel learning algorithm to train a hierarchical HAR in a unified process, simplifying the intricate training process typically linked with such architectures.

\end{itemize}

The paper is organized as follows.
Section \ref{sec:related_works} provides an overview of the related literature.
Section \ref{sec:proposed_architecture} delves into the proposed architecture and its key features.
In Section \ref{sec:evaluation}  experimental results are provided.
Finally, Section \ref{sec:conclusions} discusses the main findings of this research and addresses future research directions.

\section{Related works} \label{sec:related_works}
\subsection{Human Activity Recognition}
HAR techniques present in the literature \cite{noauthor_human_2019,bevilacqua_human_2019} can be categorized into two main approaches: vision-based HAR and sensor-based HAR. Vision-based HAR involves the analysis of images or videos captured by optical sensors, while sensor-based HAR exploits data from wearable and environmental sensors, such as accelerometers and gyroscopes \cite{minh_dang_sensor-based_2020}.
In this paper, we focus on sensor-based HAR, and the related literature will explore this specific approach.

More specifically, \cite{noauthor_human_2019} proposes an extensive study of the state-of-the-art methods for HAR together with their relative challenges. In particular, the authors divided HAR techniques into two main categories: ML-based (e.g., Support Vector Machine, K-Nearest Neighbour, and Decision Trees) and Neural Network (NN)-based (e.g., CNNs and Recurrent Neural Networks).
In \cite{khimraj_human_2020} a dataset consisting of data produced by accelerometer and gyroscope sensors of smartphones is used to show the effectiveness of different ML algorithms. This analysis showed that CNNs provide the largest recognition abilities and, for this reason, the proposed \textit{\networkNameNoSpace} is designed starting from a CNN.

Remarkably, sensor-based HAR is typically conducted over a fixed-duration window of signals and, in this perspective, the selection of the window size and sampling frequency becomes crucial when designing a HAR pipeline.
In \cite{banos_evaluating_2014} the authors extensively explore how different windowing techniques impact the recognition accuracy. The findings reveal that shorter windows, specifically those lasting 2 seconds or less, yield the highest accuracy in HAR.
In \cite{antonio_santoyo-ramon_study_2022} a study on $15$ public datasets shows that a frequency of $15 - 20$ Hz may be sufficient for HAR on a wearable device (specificity/sensitivity higher than $95\%$).
Moreover, \cite{gao_evaluation_2014} shows that increasing the sampling rate above $20$ Hz improves the recognition accuracy by just $1\%$.

One of the most promising architectures in HAR is the hierarchical approach \cite{wang_hierhar_2021}. This method enhances the discrimination between similar activities without explicitly assuming temporal relationships between actions and activities. Hierarchical HAR is a tree-based activity recognition model that, first infers the abstract activity of the user and, subsequently, identifies the specific activity in a top-down scheme  \cite{ wang_towards_2018, banos_human_2013, cho_divide_2018, pu_hrhar_2023}. The development of \textit{\networkNameNoSpace} is inspired by the hierarchical HAR solutions. This aspect will be cleared in Section \ref{sec:proposed_architecture}.

\subsection{TinyML}
The popularity of TinyML derives from its capability to process data locally on the device, improving privacy, and security, reducing latency, and enabling offline operation without a constant internet connection \cite{tiny_org}.
Most of the TinyML solutions present in the literature aim at reducing the size and complexity of the ML models \cite{buyya_is_2023,banbury_benchmarking_2021}. These solutions encompass precision scaling, which involves techniques like quantization \cite{nagel_white_2021} and model compression \cite{neill_overview_2020}, and task dropping to alleviate computational burdens \cite{buyya_is_2023}.
Another area of exploration in TinyML involves redesigning the network architecture, such as implementing approximate or dilated convolutions \cite{chollet_xception_2017,yu_multi-scale_2016}.

Regarding the on-device learning of TinyML models, several studies in the literature introduced techniques targeting either specific layers or the entire architecture of Fully Connected Neural Networks \cite{pavan_tybox_2023,ren_tinyol_2021}. Moreover, contributions focusing on the training of all the layers of Convolutional Neural Networks (CNNs) can be found in the literature \cite{craighero_-device_2023}. All these approaches focused on performing on-device learning by optimizing the memory and the computational demand, while our approach aims to enhance the on-device learning process with limited supervised data availability and adding new tasks for HAR.

\section{Proposed Methodology}\label{sec:proposed_architecture}
This section introduces the proposed methodology. 
More specifically, Section \ref{sec:problem_formulation} formalizes the proposed solution.
Section \ref{sec:overview} details the proposed \textit{\networkNameNoSpace} architecture.
Section \ref{sec:training_process}, and \ref{sec:inference_process} detail the initial off-device learning and on-device inference process of the proposed architecture, respectively.
Finally, Section \ref{sec:on-device_learning} discusses the on-device learning process.

\subsection{Formalization}\label{sec:problem_formulation}
Following the formalization introduced in \cite{wang_hierhar_2021}, the HAR multi-class classification task $T^{(0)}$ can be decomposed into $n$ sub-tasks $\{T^{(1)},\ldots,T^{(n)}\}$.
Formally, a task $T^{(i)}$ is a classifier that maps an input window of size $T$ of sequential data $\{x_{t},\ldots,x_{t-(T-1)}\}$ to its label $y^{(i)}_t$ as follows:
\[T^{(i)} \colon \{x_{t},\ldots,x_{t-(T-1)}\} \to y^{(i)}_{t}\]
being:
\begin{itemize}
    \item $x_t \in \mathbb{R}^{N_{in}}$ the input of size $N_{in}$ at time $t$, 
    \item $y^{(i)}_t$ a label belonging to the label set $\Omega^{(i)} = \{\Omega^{(i)}_1,\ldots, \Omega^{(i)}_{k_i} \}$, where $k_i$ is the total number of classes related to the task $T^{(i)}$.
\end{itemize}

We emphasize that the sequential execution of a subset of the sub-tasks set $\{T^{(1)},\ldots,T^{(n)}\}$ is equivalent to the execution of $T^{(0)}$. Indeed, in the hierarchical process, the predicted label $y^{(L)}_t$ from the last executed sub-task $T^{(L)}$ in the hierarchy belongs to the label set $\Omega^{(0)}$ of $T^{(0)}$, as follow:
\[y^{(L)}_t \in \Omega^{(0)}, \text{being} T^{(L)} \text{the last executed sub-task.}\]

\begin{figure}
    \centering
    \includegraphics[width=0.9\linewidth]{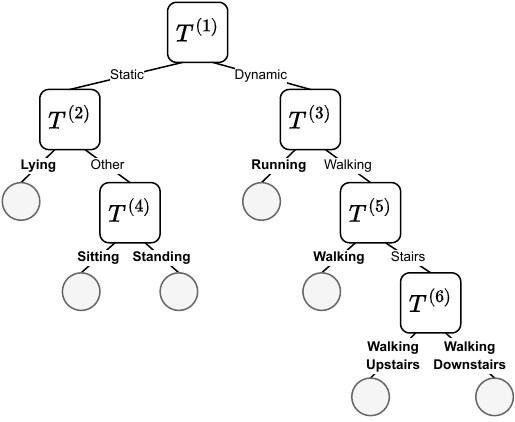}
    \caption{Example of hierarchical HAR schema.}
    \label{fig:hier-example-fig}
\end{figure}

To express the dependencies between tasks, we introduce the matrix $D$ of dimension $n\times n$, representing the dependencies among tasks as follows:
\[D =\begin{bmatrix}
 d^{(1)}_1& \ldots & d^{(1)}_n\\
\vdots & \ddots & \vdots\\
d^{(n)}_1& \ldots & d^{(n)}_n\\
\end{bmatrix} \]
where $d^{(i)}_j \in \{\Omega^{(j)} \cup \emptyset\}$ denotes whether the task $T^{(i)}$ depends on the task $T^{(j)}$.
More specifically:
\begin{itemize}
    \item $d^{(i)}_j = \emptyset$ indicates the absence of dependencies of the task $T^{(i)}$ from the task $T^{(j)}$;
    \item $d^{(i)}_j \in \Omega^{(j)}$ indicates that a dependency exists between $T^{(i)}$ and $T^{(j)}$. Moreover, the label $d^{(i)}_j$ of $T^{(j)}$ must be predicted to trigger the activation of the task $T^{(i)}$. 
\end{itemize}
In Figure \ref{fig:hier-example-fig} an example of HAR schema with $n=6$ is presented. Both the subset of the sub-tasks to be executed and their execution order are determined through a dependency structure $D$ introduced above.
As an example, to illustrate the concept of dependencies, consider task $T^{(6)}$ shown in Figure \ref{fig:hier-example-fig}. Being $\Omega^{(5)} = \{\textit{walking}, \textit{stairs}\}$, the row of $D$ associated with $T^{(6)}$ is $d^{(6)} = [\emptyset,\emptyset,\emptyset,\emptyset, \text{stairs}, \emptyset]$
indicating that, to activate $T^{(6)}$, $T^{(5)}$ must predicts \textit{stairs}.
\subsection{Proposed architecture}\label{sec:overview}
The proposed \textit{\networkNameNoSpace} architecture aims to reduce both the computational and the memory requirements of the TinyML modules for HAR (i.e., one module for each sub-task) by employing a single feature extractor (FE) module, which is used by all the tasks, and a set of multiple fully connected (FC) modules, one for each specific task. 
This approach leads to a general FE learned without specific task dependencies and FC modules specific to their respective tasks.
More formally, we refer to FE as $g(x_{t},\ldots,x_{t-(T-1)})$, and to the FC related to the task $T^{(i)}$ as $h^{(i)}(\cdot)$. Therefore, each task $T^{(i)}$ is implemented as:
\[\Tilde{y}^{(i)}_t = h^{(i)}(g(x_{t},\ldots,x_{t-(T-1)}))\]
being $\Tilde{y}^{(i)}_t$ the predicted label for task $T^{(i)}$ at time $t$.

In the proposed architecture, the function $g(\cdot)$ is implemented through a convolutional FE. 
Differently, the function $h^{(i)}(\cdot)$ of task $T^{(i)}$ is designed with $b^{(i)}$ dense layers, each having $q^{(i)}_{u_i}$ dense units with $u_i \in [1, b^{(i)}]$. The final layer of FC is a softmax dense layer that yields $k_i$ outputs (i.e., $q_{b^{(i)}}^{(i)} = k_i$).

\subsection{Off-device training process}\label{sec:training_process}

\begin{figure}
    \centering
    \includegraphics[width=0.9\linewidth]{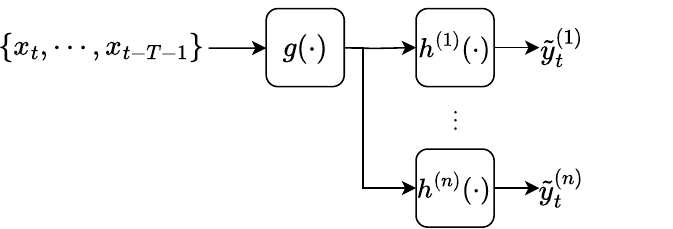}
    \caption{Overview of the off-device training process.}
    \label{fig:hisnet-overview-fig}
\end{figure}

The training process of $g(\cdot)$ and $h^{(i)}$s is jointly carried out by considering the problem as a multi-output classification problem, as shown in Figure \ref{fig:hisnet-overview-fig}. Therefore, after defining a proper loss function $L^{(i)}(y^{(i)},\tilde{y}^{(i)})$ for each sub-task $T^{(i)}$, the total loss function $L$ is defined as:
\[L=\sum_{i=1}^{n} \alpha^{(i)} L^{(i)}\text{,}\]
being $\alpha^{(i)} \in \mathbb{R}^+$ a parameter that determines how much the optimization process should focus on minimizing the loss $L^{(i)}$ associated with task $T^{(i)}$ for the given sample.
In our approach, $\alpha^{(i)}$ denotes the probability of activating the corresponding task $T^{(i)}$, which is implemented as follows:
\[\alpha^{(i)} = \begin{cases}
    1 & \text{if } d^{(i)}_j = \emptyset, \forall j\in[1,n]\text{,}\\
    \sum^{n}_{j=1} c\left(d_j^{(i)}\right) \alpha^{(j)} & \text{otherwise}
\end{cases}\]
where $c\left(d^{(i)}_j\right) \in [0,1]$ is the confidence level with which $d^{(i)}_j$ is predicted. In our implementation, the final layer of all FC modules exploits a softmax operation, hence allowing us to compute the confidence $c\left(d^{(i)}_j\right)$ as the softmax score for the corresponding label $d^{(i)}_j$. Note that $c(\emptyset)=0$.

\subsection{On-device inference process}\label{sec:inference_process}

The inference process of \textit{\networkNameNoSpace} is carried out hierarchically on the device.
Specifically, Algorithm \ref{alg:on-device-inference} reports the on-device inference process of \textit{\networkNameNoSpace}.
\begin{algorithm}
\caption{ On-device inference process}\label{alg:on-device-inference}
\begin{algorithmic}[1]
    \Procedure{Dendron\_prediction}{$x_{t},\ldots,x_{t-T-1}$}
    \State $\textit{f}_t = g(x_{t},\ldots,x_{t-(T-1)})$
    \Comment{Extract feature}
    \State $T^{(i)}$ = root task
    
    \Do
    \State $\Tilde{y}_t^{(i)} = h^{(i)}(\textit{f}_t)$
    \Comment{prediction with $T^{(i)}$}
    \State $T^{(i)} = T^{(j)}$
    \Comment{ where: $\Tilde{y}_t^{(i)} = d^{(j)}_{i}$}
    \doWhile{$\Tilde{y}_t^{(i)} \notin \Omega^{(0)}$}
    \State \textbf{return} $\Tilde{y}_t^{(i)}$
    \EndProcedure
\end{algorithmic}
\end{algorithm}
In particular, $g(\cdot)$ is executed on the input window $\{x_{t},\ldots,x_{t-(T-1)}\}$\footnote{During the inference process in HAR, input windows are usually partially overlapped. However, this does not affect the proposed solution.}.
Then, following the hierarchical schema defined in $D$, a subset of $\{h^{(1)}(\cdot),\ldots, h^{(n)}(\cdot)\}$ is sequentially executed on the output of $g(\cdot)$.
More in details, the first FC module to be executed (i.e., the root task $T^{(i)}$) is the one without any dependency, hence $d^{(i)}_j = \emptyset$, with $j \in [1,n]$.
Then, we iterate through the dependency schema $D$ until the predicted label $\Tilde{y}_t^{(i)}$ is the algorithm's final prediction (i.e., $\Tilde{y}_t^{(i)} \in \Omega^{(0)}$).



\subsection{On-Device learning process}\label{sec:on-device_learning}
In this work, our objective is to support the on-device learning of an additional task $T^{(new)}$ in limited supervised-data availability conditions. Consequently, when a user chooses to add a new task $T^{(new)}$, our objective is to minimize the amount of data labeling required to learn $T^{(new)}$.
In the proposed \textit{\networkNameNoSpace} architecture, learning a new task $T^{(new)}$ requires learning $h(\cdot)^{(new)}$, and deciding how to update the hierarchical schema $D$. Both aspects are detailed in what follows.


\subsubsection{Learn the FC module $h(\cdot)^{(new)}$}
\textit{\networkNameNoSpace} is designed in a way such that only the learning of the FC $h(\cdot)^{(new)}$ layer for $T^{(new)}$ is needed.
Consequently, learning a new task $T^{(new)}$ allows us to keep fixed $g(\cdot)$ and fine-tune the parameters of $h^{(new)}(\cdot)$ to minimize the corresponding loss $L^{(new)}$. In particular, $h^{(new)}(\cdot)$ is optimized using the standard Gradient Descent algorithm, and the memory $m_w$ required to store the weights of $h^{(new)}$ is computed as (see formalism in Section \ref{sec:overview}):
\[m_w = \sum_{u=1}^{b^{(new)}} q^{(new)}_u\times q^{(new)}_{u-1}\text{.}\]
being $b^{(new)}$ the number of dense layers of $h^{(new)}(\cdot)$, each having $q^{(new)}_{u}$ dense units with $u \in [1, b^{(new)}]$. Note that $q^{(new)}_{0}$ is the number of feature extracted by $g(\cdot)$.


\subsubsection{Update the hierarchical schema $D$}

Determining where the newly learned task $T^{(new)}$ must be integrated within the hierarchical schema established by $D$ is done as described in Algorithm \ref{alg:on-device-addition}.
\begin{algorithm}
\caption{On-device node selection process}\label{alg:on-device-addition}
\begin{algorithmic}[1]
    \Require S
    \Comment{acquired dataset for the new task $T^{(new)}$}
    \State count $= [0,\ldots,0]$
    \Comment{len(count) = num of labels of $T^{(0)}$}
    \For{$s \in S$}
    \State $\Tilde{y} = \Call{Dendron\_prediction}{s}$
    \State count[i] = count[i] + 1
    \Comment{$\Tilde{y} = \Omega^{(0)}_i$}
    \EndFor
    \State count\_tmp = \Call{Sort}{count} 
    \State $f'$ = count\_tmp[0]
    \Comment{frequency of the most predicted label}
    \State $f''$ = count\_tmp[1] 
    \Comment{frequency of the 2nd most predicted label}
    \If{$f' - f''>\delta$}
    \Comment{with $\delta \in [0,1]$}
        \State add to the node associated with $f'$
    \Else
        \State add to the nodes associated with both $f'$ and $f''$
    \EndIf
    
\end{algorithmic}
\end{algorithm}
In particular, $T^{(new)}$ is added to the most frequently predicted label if the frequency difference with the second most predicted label exceeds a threshold $\delta$. Otherwise, $T^{(new)}$ is added to both the top predicted two labels.


\begin{figure}
    \centering
    \includegraphics[width=0.7\linewidth]{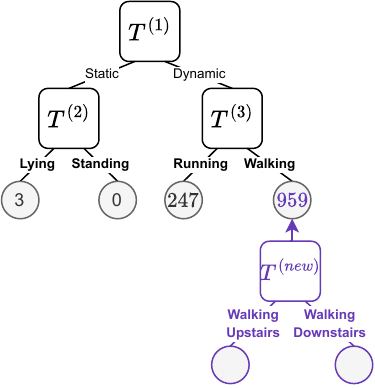}
    \caption{Example of the process for selecting the node to add a new task $T^{(new)}$.}
    \label{fig:add_new_task}
\end{figure}

Figure \ref{fig:add_new_task} reports an example of this node selection process. Specifically, we aim to add the task of distinguishing between walking upstairs and downstairs. Given that the majority of the data corresponds to walking (with $f' = \frac{959}{1209}$, and $f'' = \frac{247}{1209}$ the frequency of the two most predicted labels respectively), the walking node is considered a potential candidate for the new task $T^{(new)}$. Subsequently, if we set $\delta = 0.5$ the second heuristic is also satisfied. Indeed, the difference in frequency between the top two predicted labels is $\approx 0.59$, which exceeds $\delta = 0.5$. Note that if $\delta$ were set to $0.6$, the second heuristic would not be met, and $T^{(new)}$ would have been added to both running and walking.

\section{Experimental results} \label{sec:evaluation}

\begin{figure*}
    \centering
    \includegraphics[width=\linewidth]{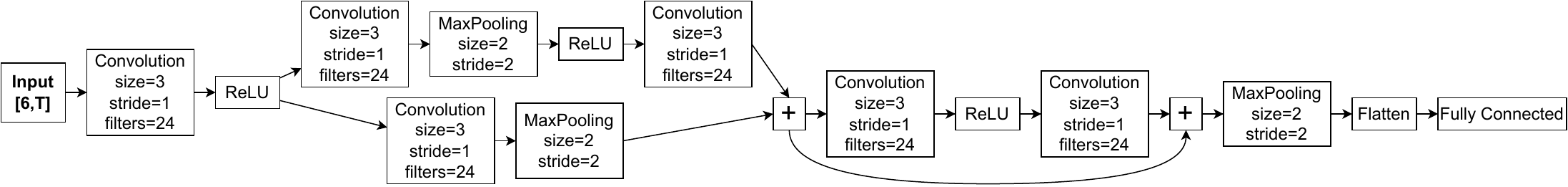}
    \caption{One-dimensional ResNetv1-6 model architecture used in the evaluation conducted in Section \ref{sec:evaluation}.}
    \label{fig:feature_extractor_uca}
\end{figure*}

This section evaluates the effectiveness and efficiency of the proposed architecture using the UCA-EHAR \cite{novac_uca-ehar_2022}, and the UCI-HAPT \cite{wermter_human_2014} datasets. 

As a comparison, the following architectures have been employed:
\begin{enumerate}
    \item \textbf{Traditional single-model solution} \cite{novac_uca-ehar_2022}: a single neural network where the output corresponds to the performed task.\label{enumerate: non_hierarchical_solution}
    \item \textbf{Hierarchical solution} \cite{wang_hierhar_2021}: multiple neural networks organized hierarchically, one for each sub-task.\label{enumerate: hierarchical_solution}
\end{enumerate}
To ensure consistency across experiments for \textit{\networkNameNoSpace}, the traditional and the hierarchical solution the evaluation employs the architecture proposed in \cite{novac_uca-ehar_2022}, a modified version of a one-dimensional ResNetv1-6 \cite{he_deep_2015}. 
Specifically, the feature extractor $g(\cdot)$ is the same as the one reported in Figure \ref{fig:feature_extractor_uca}, while the fully connected module $h^{(i)}(\cdot)$ comprises a single dense layer having the same number of neurons as the number of classes ($k_i$) to be classified. 

This section is organized as follows. 
Section \ref{sec:datasets} details the datasets used for the evaluation.
Section \ref{sec:ev_off_device} evaluates the classification capability.
Section \ref{sec:on-device_learning_ev} evaluates the performances of adding a new task during the operational life of HAR.
Finally, Section \ref{sec:ev_computation} discusses the memory usage, computation requirements, and mean latency.

\subsection{Datasets}\label{sec:datasets}

\subsubsection{UCA-EHAR dataset}
The UCA-EHAR dataset \cite{novac_uca-ehar_2022} includes gyroscopic and accelerometer data collected from smart glasses worn by $20$ people
engaged in 8 different activities.
Specifically, the dataset has been collected at $26$ Hz, and has been divided into approximately $77\%$ for training (14 subjects) and $23\%$ for testing (6 subjects) of the total samples, respectively, and, the window size has been set to $2$s as suggested in \cite{novac_uca-ehar_2022}.

\subsubsection{UCI-HAPT dataset}
The UCI-HAPT dataset \cite{wermter_human_2014} is an extension of the UCI-HAR dataset \cite{Anguita2013APD}.
It includes gyroscope and accelerometer data collected from 30 people engaged in 6 basic activities, similar to the original UCI-HAR dataset, with a waist-mounted smartphone. Additionally, the UCI-HAPT dataset introduces 6 postural transition activities, expanding on the original activity set.
Specifically, the dataset has been collected at $50$ Hz, and has been divided into approximately $70\%$ for training (21 subjects) and $30\%$ for testing (9 subjects) of the total samples, respectively, and, the window size has been set to $2.56$s as suggested in \cite{Anguita2013APD}.

\begin{figure}[b]
    \centering
    \includegraphics[width=\linewidth]{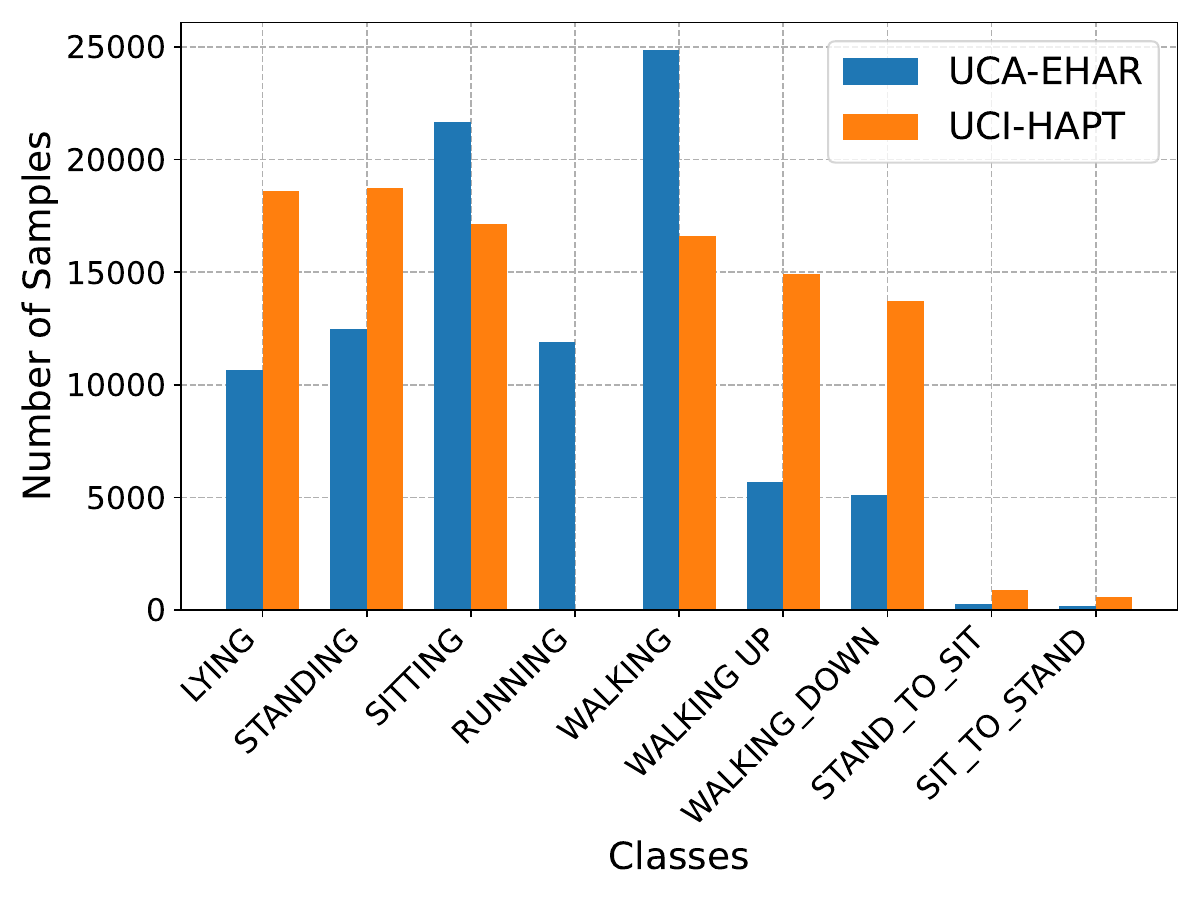}
    \caption{Number of Samples per Class for UCA-EHAR and UCI-HAPT Datasets}
    \label{fig:sample_ditribution}
\end{figure}

Figure \ref{fig:sample_ditribution} reports the distribution of samples per class for the classes used in the evaluation across both the UCA-EHAR and UCI-HAPT datasets.

\subsection{Off-device learning performance evaluation}\label{sec:ev_off_device}

In this section, we compare the classification capability of \textit{\networkNameNoSpace} with the one of the traditional and hierarchical solutions. Specifically, the following classes have been employed in this study: standing, sitting, walking, lying down, walking downstairs, walking upstairs, and running. Furthermore, for both the hierarchical solution and the proposed \textit{\networkNameNoSpace}, the employed hierarchical schema is the one reported in Figure \ref{fig:hier-example-fig}, which has been introduced in \cite{wang_towards_2018}.
The results for both datasets are summarized in Tables \ref{table:off_device_evaluation} and \ref{table:off_device_evaluation_2}. 

Table \ref{table:off_device_evaluation} compares accuracy, precision, recall, and F1-score across the three methods. Overall, \textit{\networkNameNoSpace} demonstrates the best performances, outperforming the compared approaches by at least $1-2\%$.

Table \ref{table:off_device_evaluation_2} compares the accuracy-per-class across the three methods. Specifically, the hierarchical solution generally outperforms the single-model approach, as its structure facilitates distinguishing between closely related activities. However, it struggles when classes have few supervised data, such as walking upstairs and walking downstairs for the UCA-EHAR where its performances are $4-6\%$ worse than the one of the single-model approach. On the contrary, while maintaining the hierarchical framework and having less memory usage, \textit{\networkNameNoSpace} achieves higher performance, even in cases where few supervised data are available for specific classes.

    

\begin{table}
    \caption{Off-Device classification performances comparison between the single-model solution, the hierarchical solution, and Dendron.}
    \label{table:off_device_evaluation}
    \centering
    \begin{tabular}{|ccccc|}
         \hline
         Dataset&Metric&Single-model & Hierarchical & \textbf{Dendron}\\
         \hline\hline
         &Accuracy& $0.73 \pm 0.015$& $0.75 \pm 0.011$ & $\pmb{0.77 \pm 0.012}$\\
         \textbf{UCA-}&Precision& $0.74 \pm 0.004$& $0.75 \pm 0.007$& $\pmb{0.77 \pm 0.005}$\\
         \textbf{EHAR}&Recall& $0.73 \pm 0.015$& $0.75 \pm 0.011$& $\pmb{0.77 \pm 0.012}$\\
         &F1-score& $0.72 \pm 0.009$& $0.75 \pm 0.009$& $\pmb{0.76 \pm 0.007}$\\
        
         \hline\hline
         &Accuracy& $0.90 \pm 0.008$& $0.90 \pm 0.004$ & $\pmb{0.91 \pm 0.007}$\\
         \textbf{UCI-}&Precision& $0.90 \pm 0.007$&  $0.90 \pm 0.003$ & $\pmb{0.91 \pm 0.006}$\\
         \textbf{HAPT}&Recall& $0.90 \pm 0.008$&  $0.90 \pm 0.004$ & $\pmb{0.91 \pm 0.006}$\\
         &F1-score& $0.89 \pm 0.008$&  $0.90 \pm 0.004$ & $\pmb{0.91 \pm 0.007}$\\
         \hline
    \end{tabular}
\end{table}

\begin{table}[h]
    \caption{Off-Device per-class-accuracy comparison between the single-model solution, the hierarchical solution, and Dendron.}
    \label{table:off_device_evaluation_2}
    \centering
    \begin{tabular}{|ccccc|}
         \hline
         Dataset&Class&Single-model & Hierarchical & \textbf{Dendron}\\
         \hline\hline
         &Lying& $0.99 \pm 0.022$& $0.99 \pm 0.029$& $\pmb{1.00 \pm 0.000}$\\
         &Standing& $0.24 \pm 0.069$& $0.26 \pm 0.025$& $\pmb{0.36 \pm 0.093}$\\
         \textbf{UCA-}&Sitting& $0.73 \pm 0.081$& $\pmb{0.73 \pm 0.028}$& $0.68 \pm 0.084$\\
         \textbf{EHAR}&Running& $0.84 \pm 0.031$& $0.93 \pm 0.014$& $\pmb{0.94 \pm 0.014}$\\
         &Walking& $0.80 \pm 0.061$& $\pmb{0.90 \pm 0.027}$& $0.89 \pm 0.025$\\
         &Up& $0.70 \pm 0.050$& $0.66 \pm 0.011$& $\pmb{0.70 \pm 0.024}$\\
         &Down& $\pmb{0.77 \pm 0.035}$& $0.71 \pm 0.042$& $0.75 \pm 0.031$\\

         \hline\hline
         &Lying& $0.95 \pm 0.001$ & $0.99 \pm 0.000$& $\pmb{1.00 \pm 0.000}$\\
         &Standing& $0.84 \pm 0.068$& $0.84 \pm 0.031$&$\pmb{0.85 \pm 0.029}$\\
         \textbf{UCI-}&Siting& $0.76 \pm 0.081$& $0.78 \pm 0.050$& $\pmb{0.79 \pm 0.056}$\\
         \textbf{HAPT}&Walking& $0.96 \pm 0.018$& $0.97 \pm 0.016$& $\pmb{0.98 \pm 0.005}$\\
         &Up& $0.94 \pm 0.042$ & $0.94 \pm 0.011$& $\pmb{0.95 \pm 0.024}$\\
         &Down& $0.93 \pm 0.036$& $\pmb{0.94 \pm 0.008}$& $0.91 \pm 0.026$\\
         \hline
    \end{tabular}
\end{table}

\subsection{On-device learning performance evaluation}\label{sec:on-device_learning_ev}

We now conduct an analysis to demonstrate the advantages of utilizing \textit{\networkNameNoSpace} for the on-device learning of new tasks during the operational life of HAR.
The experimental results, shown in Table \ref{tab:on-device-learning}, focus on adding two new tasks, which contain some of the classes with the least amount of data in both datasets, as indicated in Figure \ref{fig:sample_ditribution}.
These tasks are: walking downstairs vs walking upstairs, and sit-to-stand vs stand-to-sit. Specifically, two solutions are analyzed for comparison:
\begin{enumerate}
    \item \textit{\networkNameNoSpace}, using the previously learned $g(\cdot)$ and learning only $h(\cdot)^{(new)}$ and
    \item the traditional approach, where both $g(\cdot)^{(new)}$, and $h(\cdot)^{(new)}$ must be learned. This approach is employed solely for comparison purposes and is impractical for real-world applications due to its excessive memory requirements (as we will prove in Section \ref{sec:ev_computation}) and the large amount of supervised data needed.
\end{enumerate}
Moreover, to emulate the on-device incremental collection of the data, the experimental results evaluated the classification ability across various data partition percentages, starting from an initial $10\%$ and increasing incrementally by 10\% up to $100\%$.
Specifically, for the sit-to-stand vs stand-to-sit task, the UCA-EHAR dataset provides a maximum of 4 min, while the UCI-HAPT dataset offers up to 24 min. For the walking downstairs vs walking upstairs task, the UCA-EHAR dataset provides a maximum of 40 min, whereas the UCI-HAPT dataset includes up to 48 min\footnote{The UCI-HAPT dataset has $\approx 480$ minutes for walking upstairs vs. downstairs, but only 10\% was used for the evaluation to simulate supervised data scarcity condition.}.
This emulation process is necessary to assess the performance degradation relative to the amount of supervised data. We emphasize that memory is a critical constraint in TinyML applications, as one minute of gyroscopic and accelerometer data collected at $50$Hz consumes approximately $20$KB of memory.
Therefore, one of the key objectives of \textit{\networkNameNoSpace} is to enhance classification performance while minimizing the data required for learning new tasks.

\begin{table}
    \centering
    \caption{Comparison between the traditional approach and Dendron in terms of accuracy with different data percentages for the training. The tasks used for the evaluation are: walking downstairs vs walking upstairs, and sit-to-stand vs stand-to-sit.}
    \label{tab:on-device-learning}
    \begin{tabular}{|@{ }c@{ }|c|c|c|c@{ }|}
    \hline
         \multirow{3}{*}{\%}&\multicolumn{4}{|c|}{\textbf{UCA-EHAR}}\\
         \cline{2-5}
        &\multicolumn{2}{|c|}{downstairs vs upstairs} & \multicolumn{2}{|c|}{sit-to-stand vs stand-to-sit}\\
         \cline{2-5}
         &Traditional&\textbf{ Dendron} & Traditional& \textbf{Dendron}\\
         \hline
            $10$&$0.72 \pm 0.019$ & $\pmb{ 0.84 \pm 0.045 }$ & $0.68 \pm 0.057$ & $\pmb{ 0.82 \pm 0.084 }$\\
            $20$&$0.76 \pm 0.028$ & $\pmb{ 0.85 \pm 0.009 }$ & $0.74 \pm 0.062$ & $\pmb{ 0.83 \pm 0.053 }$\\
            $30$&$0.80 \pm 0.022$ & $\pmb{ 0.88 \pm 0.016 }$ & $0.76 \pm 0.054$ & $\pmb{ 0.83 \pm 0.030 }$\\
            $40$&$0.82 \pm 0.039$ & $\pmb{ 0.87 \pm 0.020 }$ &  $0.71 \pm 0.065$ & $\pmb{ 0.82 \pm 0.072 }$\\
            $50$&$\pmb{ 0.88 \pm 0.025 }$ & $0.87 \pm 0.016$ & $0.79 \pm 0.080$ & $\pmb{ 0.87 \pm 0.011 }$\\
            $60$&$\pmb{ 0.88 \pm 0.027 }$ & $0.87 \pm 0.008$ & $0.75 \pm 0.068$ & $\pmb{ 0.87 \pm 0.011 }$\\
            $70$&$\pmb{ 0.90 \pm 0.032 }$ & $0.88 \pm 0.008$ & $0.79 \pm 0.028$ & $\pmb{ 0.82 \pm 0.038 }$\\
            $80$&$0.88 \pm 0.058$ & $\pmb{ 0.89 \pm 0.003 }$ & $0.75 \pm 0.056$ & $\pmb{ 0.84 \pm 0.038 }$\\
            $90$&$0.88 \pm 0.043$ & $\pmb{ 0.89 \pm 0.010 }$ & $0.72 \pm 0.050$ & $\pmb{ 0.88 \pm 0.033 }$\\
            $100$&$\pmb{ 0.89 \pm 0.015 }$ & $\pmb{0.89 \pm 0.005}$ & $0.80 \pm 0.027$ & $\pmb{ 0.83 \pm 0.030 }$\\
          \hline
    \end{tabular}
    
    \vspace{5pt}
    
   \begin{tabular}{|@{ }c@{ }|c|c|c|c@{ }|}
    \hline
         \multirow{3}{*}{\%}&\multicolumn{4}{|c|}{\textbf{UCI-HAPT}}\\
         \cline{2-5}
        &\multicolumn{2}{|c|}{downstairs vs upstairs} & \multicolumn{2}{|c|}{sit-to-stand vs stand-to-sit}\\
         \cline{2-5}
         &Traditional& \textbf{Dendron} & Traditional& \textbf{Dendron}\\
         \hline
            $10$&$0.88 \pm 0.015$ & $\pmb{ 0.89 \pm 0.023 }$&$0.72 \pm 0.039$ & $\pmb{ 0.79 \pm 0.022 }$\\
            $20$&$0.91 \pm 0.024$ & $\pmb{ 0.93 \pm 0.007 }$&$0.77 \pm 0.026$ & $\pmb{ 0.85 \pm 0.022 }$\\
            $30$&$0.92 \pm 0.026$ & $\pmb{ 0.94 \pm 0.006 }$&$0.79 \pm 0.024$ & $\pmb{ 0.88 \pm 0.010 }$\\
            $40$&$\pmb{ 0.95 \pm 0.027 }$ & $0.94 \pm 0.007$&$0.78 \pm 0.012$ & $\pmb{ 0.86 \pm 0.030 }$\\
            $50$&$\pmb{ 0.97 \pm 0.009 }$ & $0.94 \pm 0.006$&$0.81 \pm 0.018$ & $\pmb{ 0.88 \pm 0.010 }$\\
            $60$&$\pmb{ 0.97 \pm 0.019 }$ & $0.94 \pm 0.007$&$0.81 \pm 0.014$ & $\pmb{ 0.88 \pm 0.023 }$\\
            $70$&$\pmb{ 0.98 \pm 0.009 }$ & $0.94 \pm 0.005$&$0.81 \pm 0.020$ & $\pmb{ 0.90 \pm 0.020 }$\\
            $80$&$\pmb{ 0.98 \pm 0.005 }$ & $0.95 \pm 0.006$&$0.81 \pm 0.015$ & $\pmb{ 0.88 \pm 0.018 }$\\
            $90$&$\pmb{ 0.98 \pm 0.006 }$ & $0.94 \pm 0.004$&$0.83 \pm 0.022$ & $\pmb{ 0.89 \pm 0.007 }$\\
            $100$&$\pmb{ 0.99 \pm 0.004 }$ & $0.95 \pm 0.005$&$0.84 \pm 0.023$ & $\pmb{ 0.88 \pm 0.017 }$\\            
          \hline
    \end{tabular}

\end{table}

The evaluation shows that \textit{\networkNameNoSpace} achieves higher accuracy than the traditional approach in limited supervised data availability conditions. Indeed, the traditional approach surpasses in accuracy \textit{\networkNameNoSpace} only when we use at least $40-50\%$ of the available data, and even in that case the traditional approach exhibits only an increase of $2-4\%$ in accuracy over \textit{\networkNameNoSpace}.
Furthermore, in the case of extreme supervised data-scarcity conditions, as in the case of the task of sit-to-stand vs stand-to-sit where we have only $4$ and $24$ min of data respectively for the two datasets, the traditional approach never surpasses the proposed \textit{\networkNameNoSpace} architecture.
Moreover, the experimental results prove that \textit{\networkNameNoSpace} ensures stable performances across various data availability conditions. In contrast, when \textit{\networkNameNoSpace} is not employed, the performances are highly dependent on data availability, therefore less data availability leads to a significant degradation in performance.
For instance, for the task of walking upstairs versus downstairs in the UCA-EHAR dataset when \textit{\networkNameNoSpace} is employed, the accuracy ranges from $0.84$ to $0.89$, with only a $0.05$ difference between the maximum and minimum accuracy. Conversely, when employing the traditional approach, the accuracy ranges from $0.72$ to $0.90$, resulting in a larger $0.18$ difference between the maximum and minimum accuracy.

\subsection{Computation, memory, and latency evaluation}\label{sec:ev_computation}

This section analyzes the proposed solution focusing on memory usage, computation requirements, and mean latency.
Specifically, the assessment is carried out by using the STM32-NUCLEO-F401RE evaluation board, commonly utilized in TinyML applications.
This board serves as a representative platform for evaluating the feasibility and effectiveness of the proposed architecture within the TinyML context.

\subsubsection{On-device inference evaluation}

The findings of this evaluation are summarized in Table \ref{table:comparison}. 
Specifically, The results show that our solution outperformed the existing hierarchical solution in terms of memory, computation, and mean latency. Specifically, \textit{\networkNameNoSpace} has $5\times$ less memory usage, $2\times$ less computation load, and requires $2\times$ less time per inference compared to the hierarchical solution.
Additionally, \textit{\networkNameNoSpace} exhibits only $7$ KiB increase in memory compared to the traditional single-model solution, while showing a reduced computational load compared to the traditional single-model.


\subsubsection{On-device learning evaluation}
We evaluate the memory and time required for the on-device learning of a new task $T^{(new)}$ using \textit{\networkNameNoSpace}.
We emphasize that performing on-device training using traditional approaches is not feasible in a TinyML scenario, due to the memory requirements associated with the training of the entire network.
The findings of the evaluation are summarized in Table \ref{tab:on-device}. 
Specifically, a forward-backward pass for one window sample $\{x_{t},\ldots,x_{t-(T-1)}\}$ of $2$s requires approximately $56$ms and only $8$ Bytes overhead w.r.t the inference. By storing the output of $g(\cdot)$ (the intermediate extracted feature) and performing only the forward-backward pass of $h(\cdot)$, the forward-backward pass time can be reduced to approximately $10$ms.
For instance, if a user is required to collect $30$s of data per class to learn a new binary task $T^{(new)}$ on-device, it would take approximately $1.2$s to complete one epoch over $2$s windows with 50\% overlap.

\begin{table}
   \centering 
    \caption[prova]{Comparison in terms of memory, computation, and latency using the STM32-NUCLEO-F401RE for inference.\\
    \textit{$M_{\text{FE}}^{(i)}$, $C_{\text{FE}}^{(i)}$, $M_{\text{FC}}^{(i)}$, and $C_{\text{FC}}^{(i)}$ are respectively the memory, and computation load of the $i-th$ task for the feature extractor (FE), and the fully connected (FC) modules.\footnotemark}}
    \label{table:comparison}

    \begin{tabular}{|cccc|}
    \hline
     \multirow{2}{*}{Solution}& \multirow{2}{*}{Memory}& \multirow{2}{*}{\shortstack[l]{Computation\\(MACC)}} & \multirow{2}{*}{\shortstack[l]{Mean\\Latency}}\\
     &&&\\
    \hline\hline

     \multirow{2}{*}{\shortstack[l]{Single-model}} & $M_{\text{FE}} + M_{\text{FC}}$ &$C_{\text{FE}} + C_{\text{FC}}$ & \multirow{2}{*}{$\approx47$ ms}\\
     &$\approx60$ KiB&$\approx427,242$&\\
     \hline
     \multirow{2}{*}{\shortstack[l]{Hierarchical}} & $\sum_{i}^{n} M_{\text{FE}}^{(i)} + M_{\text{FC}}^{(i)}$ &$\sum_{i} C_{\text{FE}}^{(i)} +  C_{\text{FC}}^{(i)}$ & \multirow{2}{*}{$\approx94$ ms}\\
    &$\approx 324$ KiB&$\approx851,292$&\\
    
    \hline

     \multirow{2}{*}{\textbf{Dendron}} & $\pmb{M_{\text{FE}} + \sum_{i}^{n} M_{\text{FC}}^{(i)}}$ &$\pmb{C_{\text{FE}} + \sum_{i} C_{\text{FC}}^{(i)}}$ & \multirow{2}{*}{$\pmb{\approx47 \text{ ms}}$}\\
    &$\pmb{\approx67\text{ KiB}}$&$\pmb{\approx426,444}$&\\

    \hline
    \end{tabular}
    
\end{table}

\footnotetext{The notation $\sum_{i}$ denotes a summation over a specific subset of task indexes, while $\sum_{i}^{n}$ represents the summation over all task indexes.}
\begin{table}
    \caption{Analysis of Dendron On-Device Learning of a new binary task $T^{(new)}$ for one sample of size 2s on the STM32-NUCLEO-F401RE. $g(\cdot)$ is the feature extractor, and $h(\cdot)$ is the full connected part of Dendron.}
    \label{tab:on-device}
    \centering
    \begin{tabular}{|cccc|}
    \hline
         \multicolumn{2}{|c}{} & Time & Additional Memory\\
         \hline\hline
         \multirow{2}{*}{Forward pass}
         &$g(\cdot)$& $\approx 46$ ms & 0 Bytes\\
         &$h(\cdot)$& $\approx 3$ ms & 0 Bytes\\  
         \hline\hline
         
        \multirow{2}{*}{Backward pass}
         &$g(\cdot)$& Not Required & Not Required\\
         &$h(\cdot)$& $\approx 7$ ms & 8 Bytes\\
         \hline
         
    \end{tabular}

\end{table}

\section{Conclusions and future works}\label{sec:conclusions}
This paper focused on the HAR task on wearable devices, specifically emphasizing the on-device learning of new tasks.
Specifically, \textit{\networkNameNoSpace} outperforms existing solutions in conditions of supervised data scarcity. Indeed, \textit{\networkNameNoSpace} can achieve comparable performance levels in new tasks learning on-device as other solutions, but using only 10\% of the data required by those alternative methods to achieve similarly high-performance levels. Additionally, \textit{\networkNameNoSpace} requires less memory and less computation to perform the on-device training compared to other solutions which are in general not feasible in a TinyML scenario, due to the memory requirements associated with the required training process.


Future work will encompass introducing a dynamic hierarchical schema generation procedure, implementing a sensor drift detection mechanism, and extending the on-device learning process to initialize weights based on similar tasks.


\section*{Acknowledgment}
This work was carried out in the EssilorLuxottica Smart Eyewear Lab, a Joint Research Center between EssilorLuxottica and Politecnico di Milano.


\bibliographystyle{unsrt}
\bibliography{reference}

\end{document}